\documentclass{article}
\usepackage{amsmath}
\usepackage{amssymb}
\usepackage{graphicx} % For including images
\usepackage{caption} % For better caption formatting
\usepackage{authblk}
\usepackage{color,soul}
\usepackage{array}
\usepackage{booktabs}
\usepackage{multirow}
\usepackage{hyperref}
\usepackage{subcaption}
\usepackage{caption}

\begin{document}
	
	\title{Feature-Modulated UFNO for Improved Prediction of Multiphase Flow in Porous Media}
	\date{}
	
	\author[*1]{\small Alhasan Abdellatif}
	\author[*1]{\small Hannah P. Menke}
	\author[1]{\small Florian Doster}
    \author[1]{\small Kamaljit Singh}
	\author[1]{\small Ahmed H. Elsheikh}

	\affil[1]{\footnotesize Institute of GeoEnergy Engineering (IGE), School of Energy, Geoscience, Infrastructure \& Society, Heriot-Watt University, Edinburgh, EH14 4AS, UK}

	\maketitle
	\def\thefootnote{*}\footnotetext{Corresponding authors: alhasanabdellatif@gmail.com and h.menke@hw.ac.uk}
    
    \def\thefootnote{ }\footnotetext{
    All work for this study was completed while the first author was affiliated with Heriot-Watt University}
	\begin{abstract}
        The UNet-enhanced Fourier Neural Operator (UFNO) extends the Fourier Neural Operator (FNO) by incorporating a parallel UNet pathway, enabling the retention of both high- and low-frequency components. While UFNO improves predictive accuracy over FNO, it inefficiently treats scalar inputs (e.g., temperature, injection rate) as spatially distributed fields by duplicating their values across the domain. This forces the model to process redundant constant signals within the frequency domain. Additionally, its standard loss function does not account for spatial variations in error sensitivity, limiting performance in regions of high physical importance. We introduce UFNO-FiLM, an enhanced architecture that incorporates two key innovations. First, we decouple scalar inputs from spatial features using a Feature-wise Linear Modulation (FiLM) layer, allowing the model to modulate spatial feature maps without introducing constant signals into the Fourier transform. Second, we employ a spatially weighted loss function that prioritizes learning in critical regions. Our experiments on subsurface multiphase flow demonstrate a 21\% reduction in gas saturation Mean Absolute Error (MAE) compared to UFNO, highlighting the effectiveness of our approach in improving predictive accuracy.
	\end{abstract}

\section{Introduction}

%Simulation of complex physical processes, such as subsurface multi-phase flow, is critical for effective management of resources in applications ranging from hydrocarbon production to groundwater remediation. 
Accurate simulation of the movement of fluids in porous media underpins strategies for sustainable storage \& extraction \cite{Zhang2022_CO2ReactiveFlow_GeoEnergy}, environmental monitoring, and contaminant transport \cite{Thom2023_ContaminantTransport_VadoseZone}, making these simulations indispensable in both industry and research. Traditionally, the simulation of subsurface fluid dynamics has relied on numerical solvers based on implicit formulations of partial differential equations (PDEs) \cite{orr2007theory,mohamed2010reservoir,dogru2011new}. These solvers discretize the computational domain into grids and apply finite difference, finite element, or finite volume methods to approximate the governing equations.

While such methods have been successful in reproducing the evolution of pressure, saturation, and other key variables over time, they encounter significant challenges when applied to large, high-resolution grids. The computational burden increases dramatically with grid refinement, making real-time decision-making and dynamic calibration (history matching) increasingly difficult in practical field applications \cite{doughty2010investigation}.

In recent years, data-driven surrogate models have emerged as a promising alternative to traditional solvers \cite{zhu2018bayesian,zhong2019predicting,wang2021physics,wen2021ccsnet}. These models leverage machine learning techniques to learn the underlying mappings directly from data, potentially reducing the computational costs associated with high-fidelity numerical simulations. Among these approaches, the Fourier Neural Operator (FNO) has demonstrated notable advantages by learning mappings in the frequency domain \cite{li2020fourier}. By performing a Fourier transform on the input fields, FNO is able to parameterize the low-frequency components of the solution, thereby achieving inherent resolution independence. However, while FNO efficiently captures global system behaviors, it struggles to accurately reproduce high-frequency details that are critical for modeling localized phenomena.

To overcome this limitation, the UNet-enhanced Fourier Neural Operator (UFNO) was developed by integrating a parallel UNet branch that is specifically designed to preserve high-frequency information \cite{wen2022u}. This hybrid architecture combines the strengths of both FNO and UNet to provide a more comprehensive representation of the underlying dynamics. However, in the UFNO paper, scalar control inputs, such as injection rates or temperature, are treated by spatially duplicating them. This treatment forces the model to process constant signals in the frequency domain, which is not the most effective approach as this results in redundant computations and suboptimal learning of the underlying physics. Moreover, often the loss function applies a uniform penalty across the spatial domain, ignoring local variations in error sensitivity. This uniform approach may compromise model performance in areas where even small errors carry significant physical implications.

In our work, we introduce UFNO-FiLM, an enhanced surrogate model that addresses these limitations through two key innovations. First, we decouple scalar inputs from spatial features by incorporating a Feature-wise Linear Modulation (FiLM) layer \cite{perez2018film}. Rather than spatially duplicating scalar values, the FiLM layer processes these inputs via a learnable module that generates channel-specific modulation parameters. These parameters are then used to affine-transform the spatial feature maps, ensuring that constant scalar signals modulate the learned frequency-domain representations effectively without diluting critical spatial features. Second, we adopt a spatially weighted loss function during training, which assigns greater importance to regions of the domain that are crucial for capturing complex flow dynamics. By emphasizing errors in these key areas, the adaptive loss strategy guides the network to produce more accurate overall predictions.

The remainder of this paper is organized as follows. In Section \ref{LITREV}, we review the related work in subsurface simulation and data-driven surrogate modeling. Section \ref{METHOD} details the UFNO-FiLM architecture and our proposed method. Section \ref{exp} presents a comprehensive evaluation of our model, and Section \ref{conclusion} concludes with a discussion of the results and potential directions for future research.

\section{Related Work}
\label{LITREV}
Surrogate modeling for subsurface flow has attracted significant attention over the past decades. In this section, we review prior work on both classical, physics-based surrogate models and modern data-driven approaches. 

\paragraph{Classical physics-based surrogate models}
Traditional approaches for simulating subsurface flow rely on solving the governing nonlinear partial differential equations using techniques such as finite difference, finite element, or finite volume methods \cite{dogru2011new}. However, the high computational cost associated with fine-grid discretization and multiscale heterogeneity has motivated the development of reduced-order models (ROMs). Methods such as coarse-grid modeling and proper orthogonal decomposition (POD)-based ROM reduce the complexity by mapping full simulation trajectories into a low-dimensional space \cite{cardoso2009development, he2013reduced, he2014reduced, jin2018reduced, van2006reduced, xiao2019non, yang2016fast}. An alternative to analysing the state vectors and reducing the complexity of the problem with proper orthogonal decomposition methods is to simplify the physics of the problem to be solved beforehand. Vertically integrated modelling \cite{ de2024determining, bandilla2019guideline, nilsen2016fully, nordbotten2011geological, class2009benchmark} is a powerful tool for geological carbon storage for general subsurface flow phenomena.  While these techniques can substantially speed up computations, they often simplify critical aspects of the flow physics, which may lead to loss of fidelity in representing sharp interfaces and localized heterogeneities.

\paragraph{Physics-informed and data-driven approaches}
Recent years have witnessed a surge in physics-informed machine learning methods aimed at integrating data-driven approaches with physical laws. For example, Fraces et al. \cite{fraces2020physics} employ physics-informed neural networks (PINNs) combined with transfer learning and generative techniques to tackle two-phase immiscible transport problems. Similarly, Cai et al. \cite{cai2022physics} provide a comprehensive review of PINNs for inverse problems in fluid mechanics, although Fuks and Tchelepi \cite{fukes2020limitations} point out that PINN formulations may struggle to capture sharp saturation fronts inherent to nonlinear subsurface flow dynamics.

\paragraph{Data-driven surrogate modeling}
Data-driven surrogate models bypass the need for explicit solution of the underlying PDEs by directly learning the mapping from input parameters (e.g., permeability distributions, injection rates) to simulation outputs. Early approaches include the use of Gaussian processes for history matching in unconventional reservoirs \cite{hamdi2017gaussian} and polynomial chaos expansions for modeling channelized reservoirs \cite{bazargan2015surrogate}. Feed-forward neural networks have been applied for history matching \cite{costa2014application}, while convolutional architectures such as U-Net \cite{cicek20163d} and deep convolutional encoder–decoder networks \cite{zhu2018bayesian} have demonstrated promising results in capturing spatially distributed flow fields. Recurrent architectures like R-U-Net \cite{tang2020deep} have further advanced the capability to predict dynamic behaviors over multiple time steps. Recent advances in neural operators, particularly Fourier Neural Operators (FNOs) \cite{li2020fourier}, have shown remarkable capabilities in learning solution operators for parametric PDEs while maintaining resolution independence. The U-Net-enhanced Fourier Neural Operator (UFNO) \cite{wen2022u} further combines Fourier-based global modeling with U-Net's local feature extraction, demonstrating improved accuracy for subsurface flow predictions.

Building on these developments, our work introduces a novel surrogate model that integrates the strengths of Fourier-based neural operators with modern deep learning techniques. Inspired by the success of the UNet-enhanced Fourier Neural Operator (UFNO) in capturing global system behavior \cite{wen2022u}, we further enhance the architecture by incorporating Feature-wise Linear Modulation (FiLM) layers. This design decouples scalar control inputs from spatial features, thereby reducing redundant computations and enabling more effective learning of both global dynamics and localized high-frequency details.

\section{Methodology}
\label{METHOD}
In this section, we present the methodology behind our proposed UFNO-FiLM model, which extends the UNet-enhanced Fourier Neural Operator (UFNO) through two key innovations. First, we integrate a Feature-wise Linear Modulation (FiLM) layer to decouple scalar inputs from spatial features. Second, we introduce a spatially weighted loss function to prioritize error reduction in physically critical regions. Together, these modifications enable the model to more effectively capture the underlying physical processes and enhance overall predictive accuracy.

In the original UFNO framework, scalar inputs (e.g., temperature, injection rate) are spatially duplicated to match the dimensionality of spatial fields, inadvertently forcing the network to propagate constant signals through the Fourier layers. This results in unnecessary frequency-domain computation and can dilute the learning of meaningful spatial patterns. To overcome this limitation, UFNO-FiLM incorporates a FiLM layer \cite{perez2018film} that processes scalar inputs independently through a small learnable module and uses the resulting modulation parameters to condition the spatial feature maps. In addition, we introduce a spatially weighted loss function that assigns greater importance to regions known—based on physical insight—to be more sensitive or dynamically active. The following subsections describe these two components in detail. 

\subsection*{Feature-wise Linear Modulation (FiLM) Layer}
\label{subsec:film}

\begin{figure}[h]
\centering
\includegraphics[width=\textwidth]{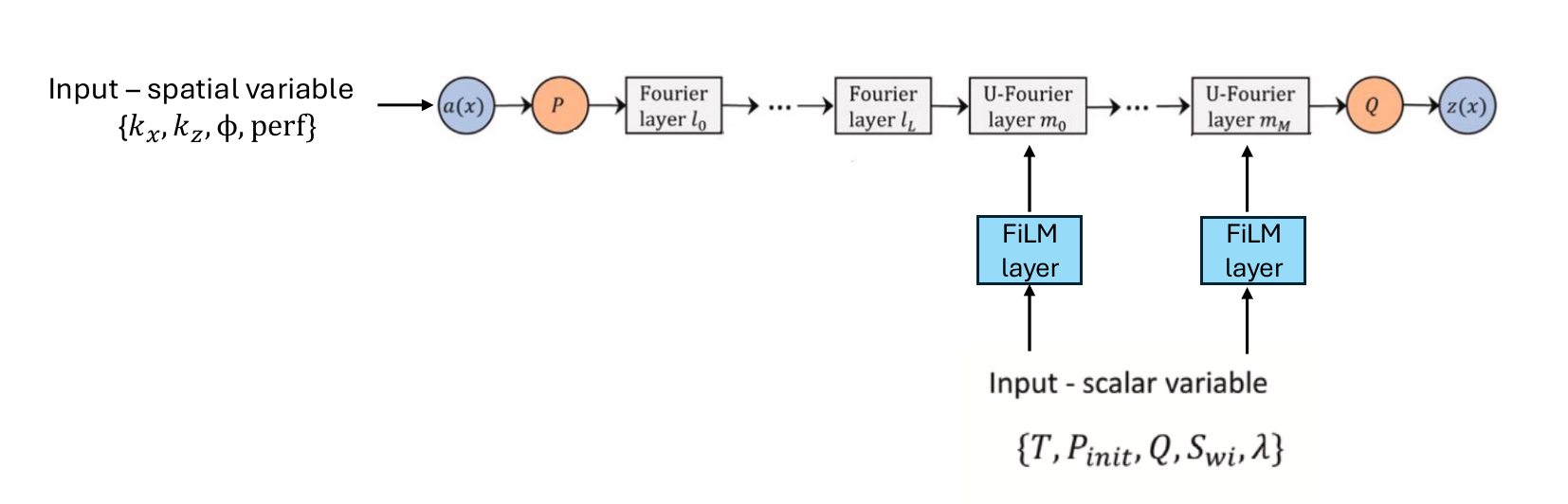}  % Adjust the filename if needed
\caption{Architecture of the UFNO-FiLM model. The FiLM layer modulates spatial feature maps based on scalar inputs, while the spatially weighted loss improves predictive accuracy.}
\label{fig:ufno_film}
\end{figure}

The FiLM layer is a mechanism originally introduced to perform feature-wise affine transformations on neural activations \cite{perez2018film}. In our UFNO-FiLM model, the FiLM layer serves the purpose of integrating scalar inputs into the network without compromising the inherent spatial characteristics learned by the main UFNO branch, as illustrated in Figure~\ref{fig:ufno_film}.

Let us denote the spatial feature maps extracted from the UFNO branch as \( F \in \mathbb{R}^{H \times W \times C} \), where \( H \) and \( W \) are the height and width of the feature maps, and \( C \) is the number of channels. Each channel represents a distinct feature extracted by the network, encoding different spatial patterns or activations. The scalar inputs, represented by a vector \( \mathbf{s} \in \mathbb{R}^{d} \) (where \( d \) is the number of scalar features), are processed through a learnable transformation \( g \) that produces modulation parameters:
\[
\left( \gamma, \beta \right) = g(\mathbf{s}), \quad \gamma, \beta \in \mathbb{R}^{C}.
\]
Here, $g: \mathbb{R}^{d} \to \mathbb{R}^{C}$ is implemented as a multi-layer perceptron (MLP) and optimized jointly with the rest of the network.  The role of the FiLM layer is to modulate each channel of the spatial features using the parameters \( \gamma \) and \( \beta \) via a channel-wise affine transformation:
\[
\widetilde{F}_{i,j,c} = \gamma_c \cdot F_{i,j,c} + \beta_c, \quad \forall \, i \in \{1,\dots,H\}, \, j \in \{1,\dots,W\}, \, c \in \{1,\dots,C\}.
\]

This operation effectively scales and shifts the spatial features according to the scalar input information. By isolating the scalar inputs from the Fourier domain transformations, the FiLM layer prevents constant signals (which have no spatial variation) from unnecessarily influencing the learned frequency representations. Instead, these inputs modulate the network’s activations in a controlled manner, enhancing the network’s expressiveness and capacity to model complex spatial-temporal dynamics.

\subsection*{Spatially Weighted Loss Function}
\label{subsec:weightedloss}

A critical challenge in training neural operators for physical simulations is ensuring that the model pays adequate attention to regions of the domain that are of higher physical importance. In conventional loss functions, such as the Mean Squared Error (MSE), each spatial location contributes equally to the overall error, which may not be ideal when certain areas exhibit more significant physical variations or are more critical for the overall system performance.

To address this issue, we propose a spatially weighted loss function. For a given sample \( n \), let \( y_{n}(i,j) \) denote the ground truth at spatial location \( (i,j) \) and \( \widehat{y}_{n}(i,j) \) denote the corresponding prediction. We assign a weight \( w(i,j) \) to each spatial location, where the weights are determined by prior domain knowledge. The spatially weighted MSE loss is then defined as:
\[
\mathcal{L}_{\text{SW}} = \frac{1}{N} \sum_{n=1}^{N} \sum_{i=1}^{H} \sum_{j=1}^{W} w(i,j) \left( y_{n}(i,j) - \widehat{y}_{n}(i,j) \right)^2,
\]
where \( N \) is the total number of training samples. The weighting function \( w(i,j) \) is designed to assign higher values to regions where accurate predictions are critical (e.g., areas with steep gradients or significant physical transitions) and lower values to regions where errors are less consequential.

In practice, the weight map \( w \) can be derived from a variety of sources. For example, if physical simulations indicate that certain spatial regions are associated with higher fluxes or stress concentrations, these regions can be manually assigned higher weights. Alternatively, an adaptive weighting strategy can be implemented where the weights are iteratively updated based on the model’s performance during training. In our experiments, the weight map \( w \)  is computed as the per-pixel mean of the training images, which captures the statistical distribution of plume migration across the dataset. This approach emphasizes the regions where plume migration is most prominent, ensuring that these critical areas are assigned higher importance during model training. The resulting weight map is then fixed and applied uniformly to all training samples.

\section{Experiments and Results}
\label{exp}

In this section, we present a comprehensive evaluation of our proposed UFNO-FiLM model on the subsurface simulation dataset used in \cite{wen2022u}. 

In this section, we present a comprehensive evaluation of the proposed UFNO-FiLM model using the benchmark multiphase-flow dataset introduced in the original UFNO studies~\cite{wen2022u}. We compare UFNO-FiLM against baseline models across multiple metrics and scenarios to demonstrate the impact of FiLM-based conditioning and spatially weighted loss on predictive accuracy.

\paragraph{Dataset Description}

We evaluate our model using the same dataset introduced in the original UFNO work~\cite{wen2022u}. The dataset consists of 5000 numerical simulations of two-phase (CO$_2$--water) flow in a radially symmetric reservoir, generated using the ECLIPSE E300 compositional simulator. Each simulation models 30 years of CO$_2$ injection under varying geological and operational conditions, including heterogeneous permeability and porosity fields, anisotropy ratios, multiple injection rates, and reservoir thicknesses ranging from 12~m to 200~m. For each realization, the dataset provides 24 temporal snapshots of the evolving gas saturation and pressure buildup fields defined on a two-dimensional radial grid.

Following the experimental protocol of~\cite{wen2022u}, the dataset is divided into 4000 samples for training, 500 for validation, and 500 for testing. Consistent with the preprocessing in the UFNO framework, gas saturation fields are kept in their physical range $[0,1]$, whereas the pressure buildup fields are standardized to zero mean and unit variance prior to training. In addition, each simulation includes an active-cell mask that identifies valid reservoir regions; all loss computations are performed only over these active cells, ensuring that padded or inactive regions do not contribute to the error.

The original UFNO loss formulation employs a relative $L_p$ loss applied both to the field values and to their first radial derivatives, which helps capture sharp saturation fronts and pressure gradients near the injection well. Because our work focuses on modeling gas saturation evolution rather than pressure transients, we do not include pressure-field predictions in our main results. Nonetheless, we follow the same normalization and masking strategy for consistency and comparability with prior work.

\begin{figure}[h]
    \centering
    \begin{subfigure}[b]{0.48\textwidth}
        \centering
        \includegraphics[width=\textwidth]{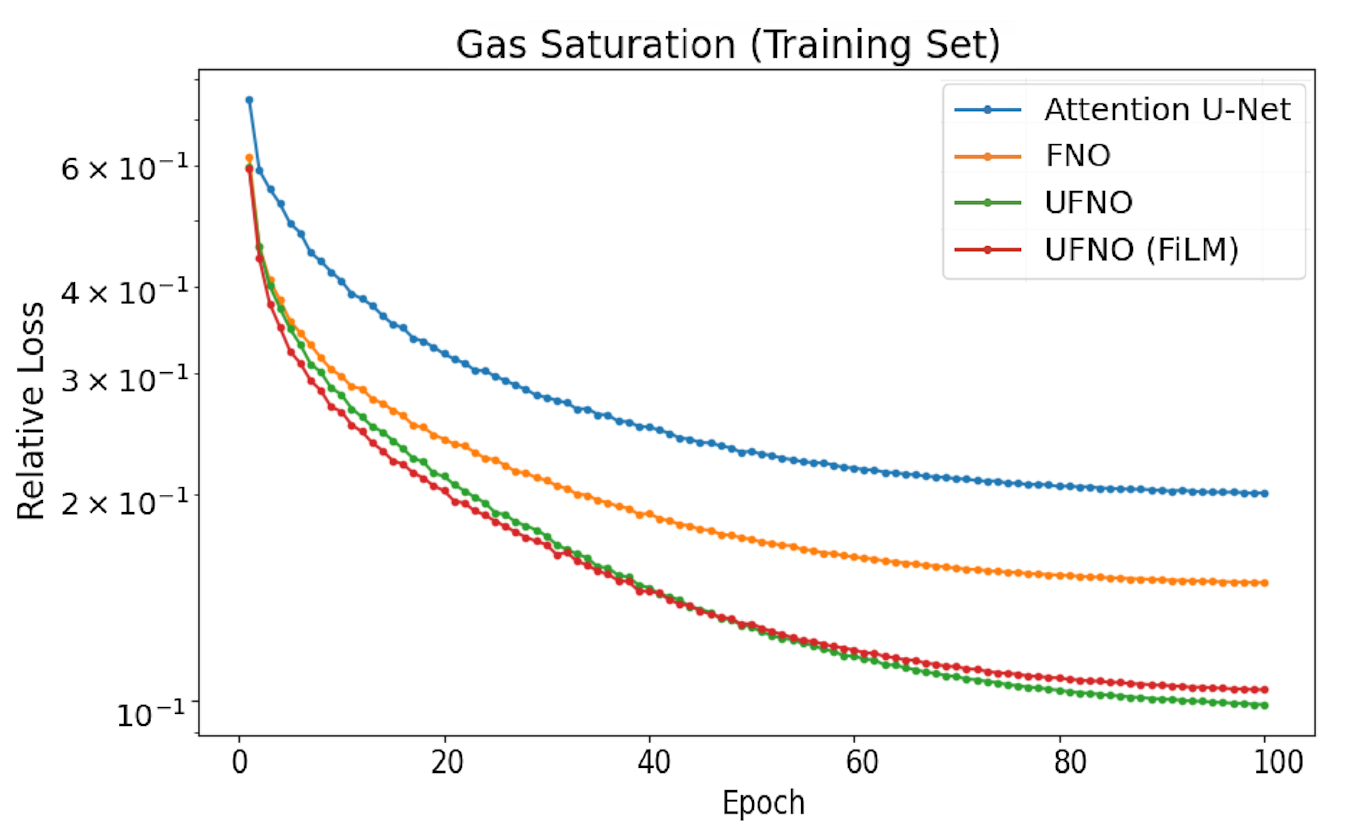} % Replace with your training curves file
        \caption{Training curves}
        \label{fig:training_curves}
    \end{subfigure}
    \hfill
    \begin{subfigure}[b]{0.48\textwidth}
        \centering
        \includegraphics[width=\textwidth]{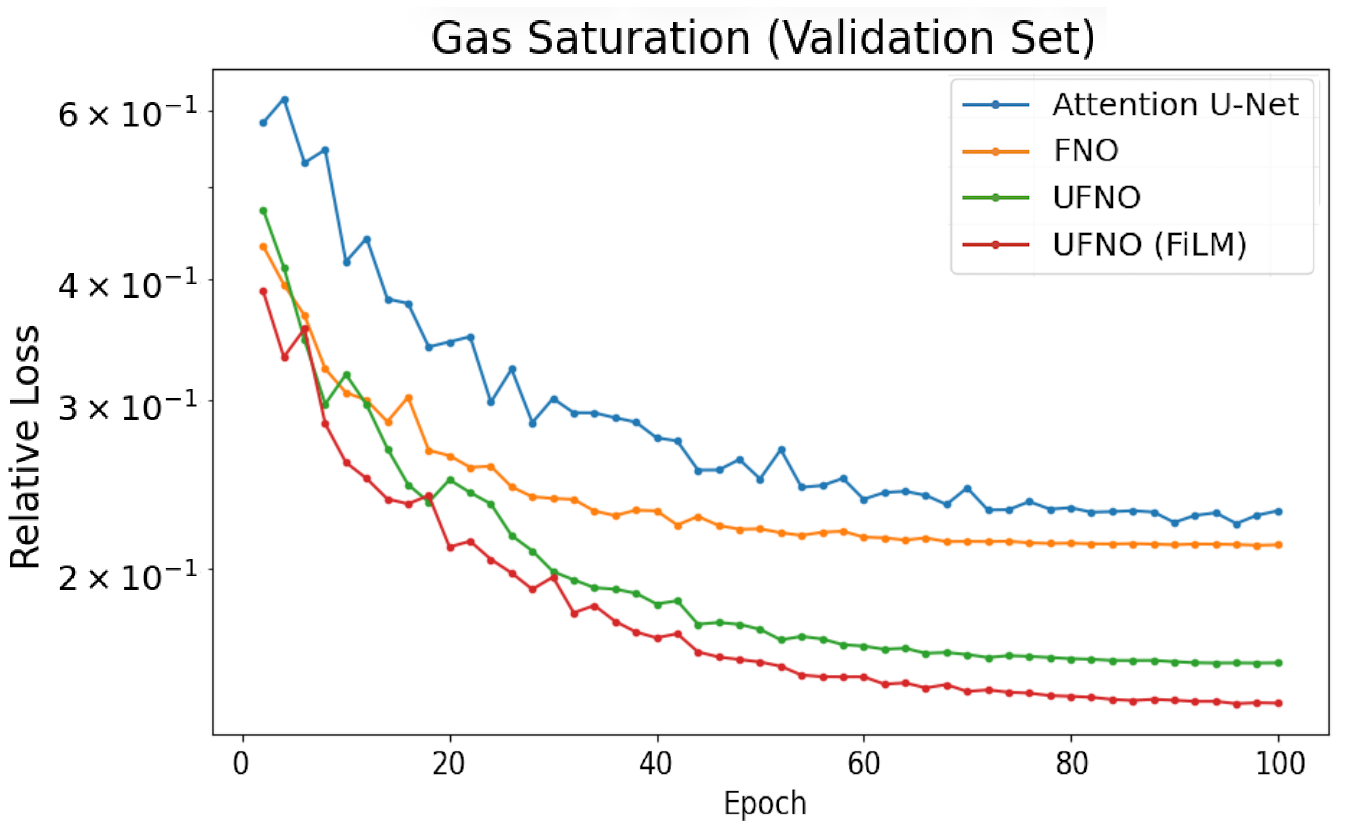} % Replace with your validation curves file
        \caption{Validation curves}
        \label{fig:validation_curves}
    \end{subfigure}
    \caption{Comparison of training and validation curves for Attention UNet, FNO, UFNO, and UFNO-FiLM. (a) Training curves show the loss decreases over epochs. (b) Validation curves highlight that UFNO-FiLM achieves the lowest validation error.}
    \label{fig:curves_side_by_side}
\end{figure}

\paragraph{Training and Validation Curves.} 
Figures~\ref{fig:training_curves} and~\ref{fig:validation_curves} show the training and validation loss trajectories for the Attention UNet, FNO, UFNO, and UFNO-FiLM models. All models exhibit stable convergence during training, with their losses decreasing monotonically over the optimization epochs. While UFNO-FiLM displays a training-loss trend comparable to that of UFNO, it achieves the lowest validation error among all models. This superior validation performance indicates that the FiLM-based conditioning and spatially weighted loss enhance the model’s generalization capabilities and mitigate overfitting relative to the baseline architectures.

\begin{figure}[h]
    \centering
    \begin{subfigure}[b]{0.48\textwidth}
        \centering
        \includegraphics[width=\textwidth]{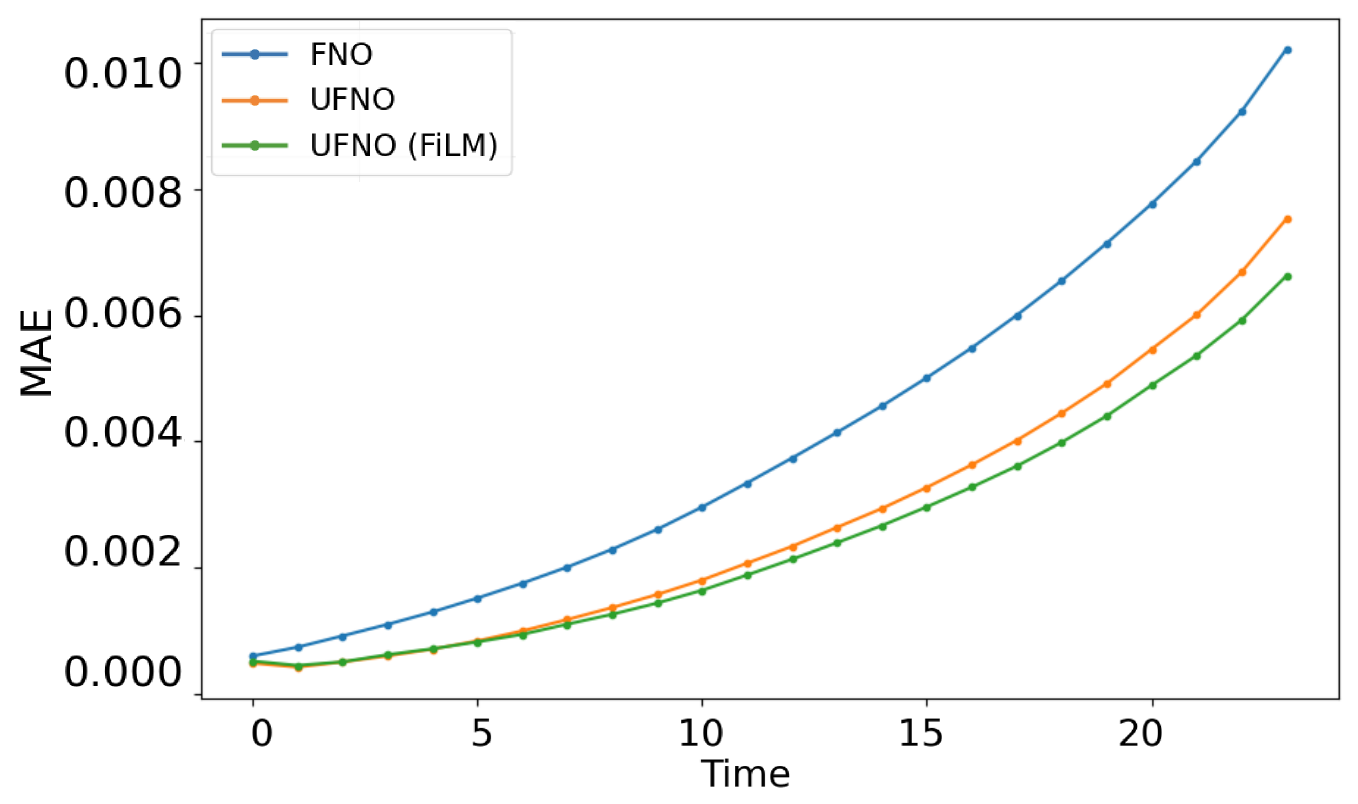} % Replace with your MAE figure filename
        \caption{MAE Comparison}
        \label{fig:mae_comparison}
    \end{subfigure}
    \hfill
    \begin{subfigure}[b]{0.48\textwidth}
        \centering
        \includegraphics[width=\textwidth]{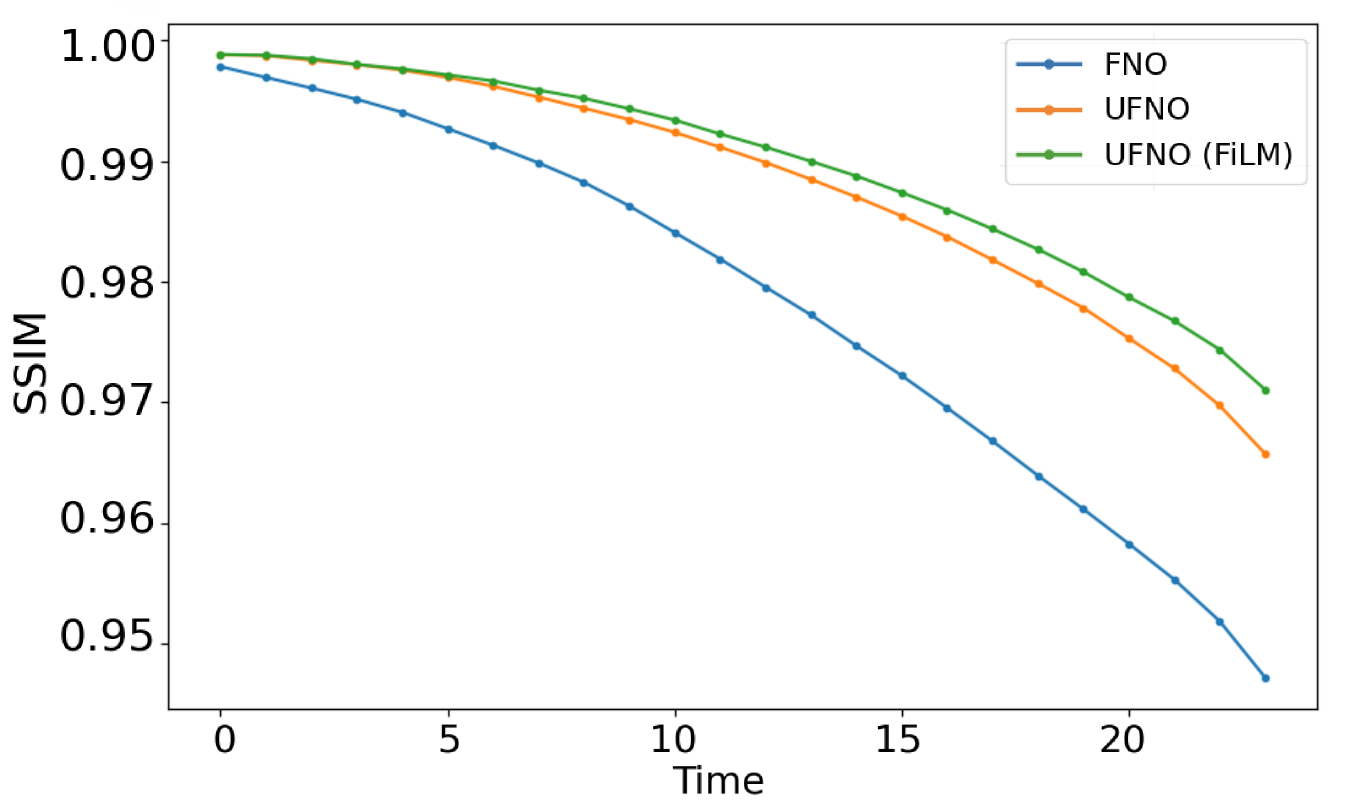} % Replace with your SSIM figure filename
        \caption{SSIM Comparison}
        \label{fig:ssim_comparison}
    \end{subfigure}
    \caption{Comparison of MAE and SSIM between predictions and ground truth at different injection times for FNO, UFNO, and UFNO-FiLM. The results show that UFNO-FiLM outperforms the other models by achieving lower MAE and higher SSIM values.}
    \label{fig:comparison_side_by_side}
\end{figure}

\paragraph{Performance at different injection times.} 
Figure~\ref{fig:comparison_side_by_side} presents a comparison of the Mean Absolute Error (MAE) and the Structural Similarity Index Measure (SSIM) computed between the predictions and ground truth at various injection times. The left subfigure shows the MAE comparison for FNO, UFNO, and UFNO-FiLM, where UFNO-FiLM consistently achieves lower error values. The right subfigure displays the SSIM comparison, where higher SSIM values for UFNO-FiLM indicate improved preservation of spatial structures. Together, these results demonstrate that incorporating the FiLM layer into the UFNO framework enhances predictive accuracy and enhances the model’s ability to reproduce fine-scale spatial features.

\paragraph{Visual Comparison of Model Predictions.}

\begin{figure}[h]
    \centering
    \includegraphics[width=\textwidth]{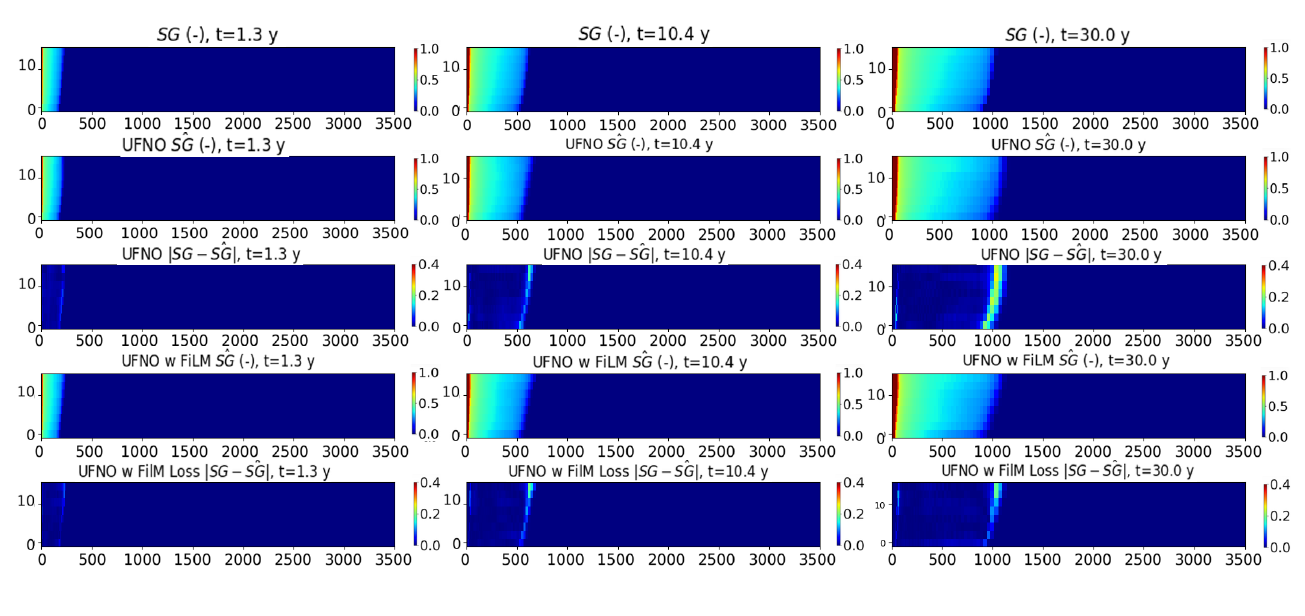} 
    \caption{Qualitative comparison of UFNO and UFNO-FiLM predictions at several injection times. Each block shows the ground truth, model predictions, and associated error maps. UFNO-FiLM exhibits visibly reduced errors and improved reconstruction of plume structure.}
    \label{fig:film_improvement}
\end{figure}

\begin{figure}[h]
    \centering
    \includegraphics[width=\textwidth]{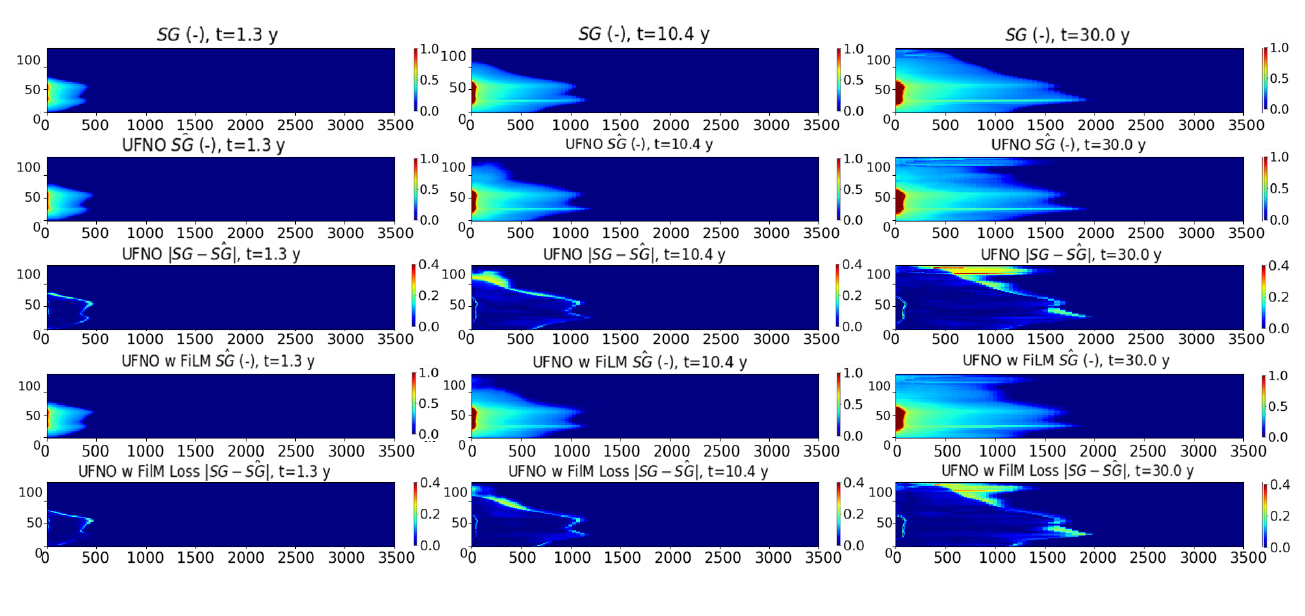}
    \caption{Additional qualitative comparison on a different test sample. UFNO-FiLM again shows improved predictive accuracy and reduced spatial errors compared with UFNO.}

    \label{fig:film_improvement2}
\end{figure}

Figures~\ref{fig:film_improvement} and~\ref{fig:film_improvement2} provide a visual overview of model predictions across different injection times. Both figures are organized into five rows representing, from top to bottom: the ground truth, UFNO prediction, UFNO error, UFNO-FiLM prediction, and UFNO-FiLM error. In these figures, the error maps clearly illustrate that the UFNO-FiLM model consistently reduces spatial errors compared to the standard UFNO approach.

\paragraph{Visual Comparison of Model Predictions.}

Figures~\ref{fig:film_improvement} and~\ref{fig:film_improvement2} provide qualitative comparisons between UFNO and UFNO-FiLM across multiple injection times. Each figure displays the ground truth saturation fields alongside the predictions and corresponding error maps for both models. The visualizations show that UFNO-FiLM consistently produces lower-magnitude and more spatially localized errors, indicating a clearer recovery of plume morphology and improved fidelity relative to UFNO.

\paragraph{Ablation Study.} 
Table~\ref{tab:metrics} summarizes the performance of various models on the validation set. Our results indicate that the incorporation of the FiLM layer and spatially weighted loss function yields consistent improvements across all metrics. In particular, the combination of the FiLM layer with the weighted loss (UFNO-FiLM \& WLoss) achieves the best MAE (0.0022) and SSIM (0.9904), while the UFNO-FiLM model alone attains the lowest Relative Error (0.1451). These improvements highlight the effectiveness of decoupling scalar inputs from spatial features and adopting a loss function that accounts for local error sensitivity.

Table~\ref{tab:metrics} summarizes the validation performance of all models. The results show that both components of our approach, the FiLM-based conditioning and the spatially weighted loss, contribute to consistent improvements across the evaluated metrics. The combined model (UFNO-FiLM~\&~WLoss) attains the lowest MAE (0.0022) and second best SSIM (0.9904), while UFNO-FiLM alone achieves the lowest relative error (0.1451). These findings demonstrate the benefit of decoupling scalar inputs from spatial features and incorporating spatially informed weighting to better capture localized flow dynamics.

\begin{table}[h]
\centering
\caption{Comparison of metrics across models on the validation set. The best value is in \textbf{bold} and the second best is \underline{underlined}.}
\label{tab:metrics}
\resizebox{1\textwidth}{!}{%
\begin{tabular}{|l|c|c|c|c|c|}
\toprule
Metric/Model          & FNO    & UFNO   & UFNO-FiLM   & UFNO-WLoss  & UFNO-FiLM \& WLoss \\
\midrule
\textbf{MAE}      & 0.0041 & 0.0028 & 0.0025   & \underline{0.0023}   & \textbf{0.0022} \\
\textbf{SSIM}    & 0.9780 & 0.9879 & 0.9895   & \textbf{0.9906}      & \underline{0.9904} \\
\textbf{Relative Error} & 0.2120 & 0.1598 & \textbf{0.1451}   & 0.1583      & \underline{0.1510} \\
\bottomrule
\end{tabular}%
}
\end{table}

\paragraph{Training and validation performance with $10\%$ data}

\begin{figure}[h]
    \centering
    \begin{subfigure}[b]{0.48\textwidth}
        \centering
        \includegraphics[width=\textwidth]{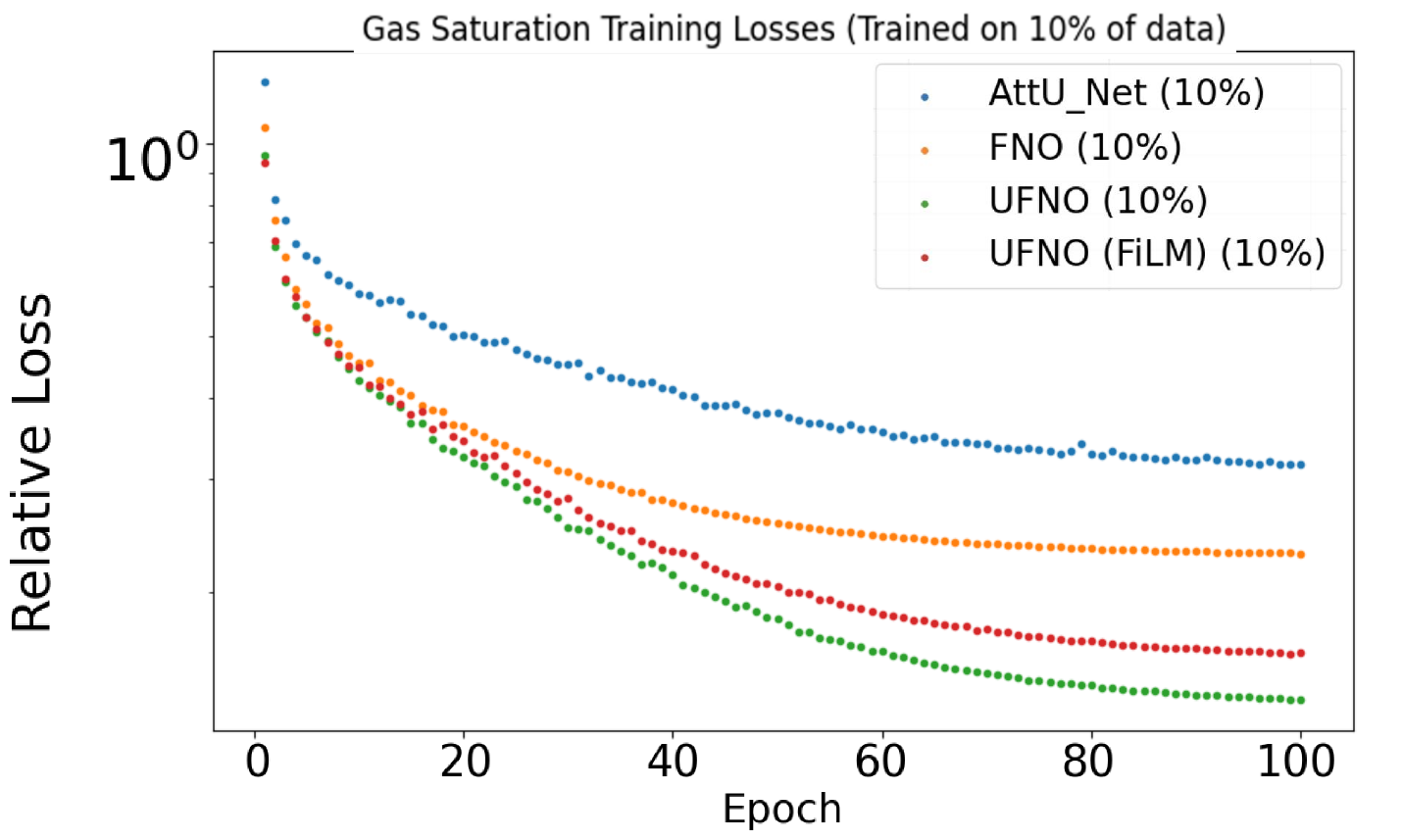} 
        \caption{Training curves}
        \label{fig:training_curves_10p}
    \end{subfigure}
    \hfill
    \begin{subfigure}[b]{0.48\textwidth}
        \centering
        \includegraphics[width=\textwidth]{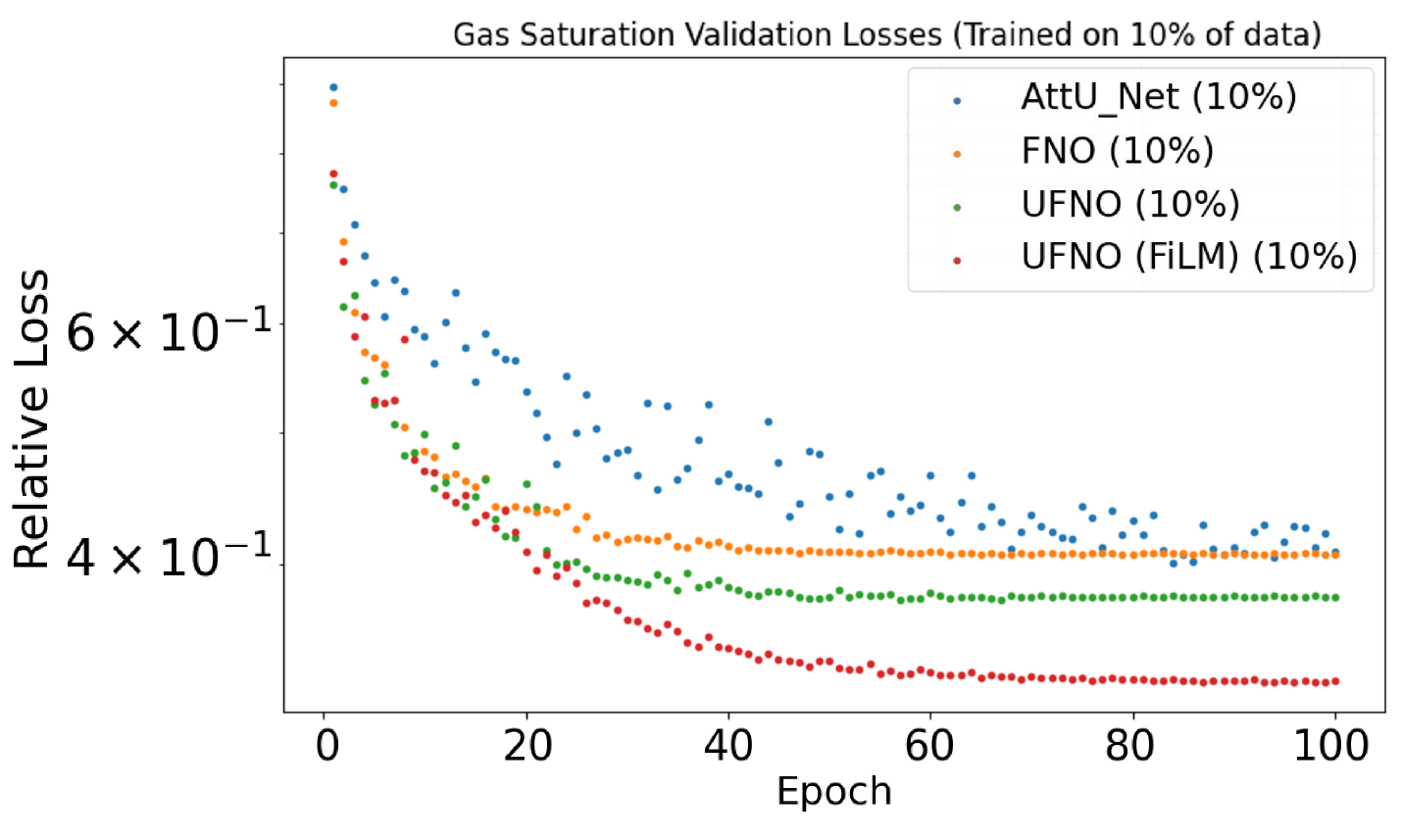} 
        \caption{Validation curves}
        \label{fig:validation_curves_10p}
    \end{subfigure}
    \caption{Comparison of training and validation curves for UFNO, and UFNO-FiLM on the gas saturation problem.}
    \label{fig:curves_side_by_side_10p}
\end{figure}

Figure~\ref{fig:curves_side_by_side_10p} present the training and validation loss curves for models trained using only $10\%$ of the available data. The validation performance highlights a notable advantage of the UFNO-FiLM model, which achieves the lowest validation error among all models. This suggests that the FiLM-based adaptation enhances the model's generalization capabilities when data availability is limited. The improved performance in this data-scarce regime indicates that UFNO-FiLM can effectively capture relevant patterns with fewer training examples, making it a promising approach for scenarios where acquiring large-scale training data is challenging.

\begin{figure}[h]
    \centering
    \begin{subfigure}[b]{0.48\textwidth}
        \centering
        \includegraphics[width=\textwidth]{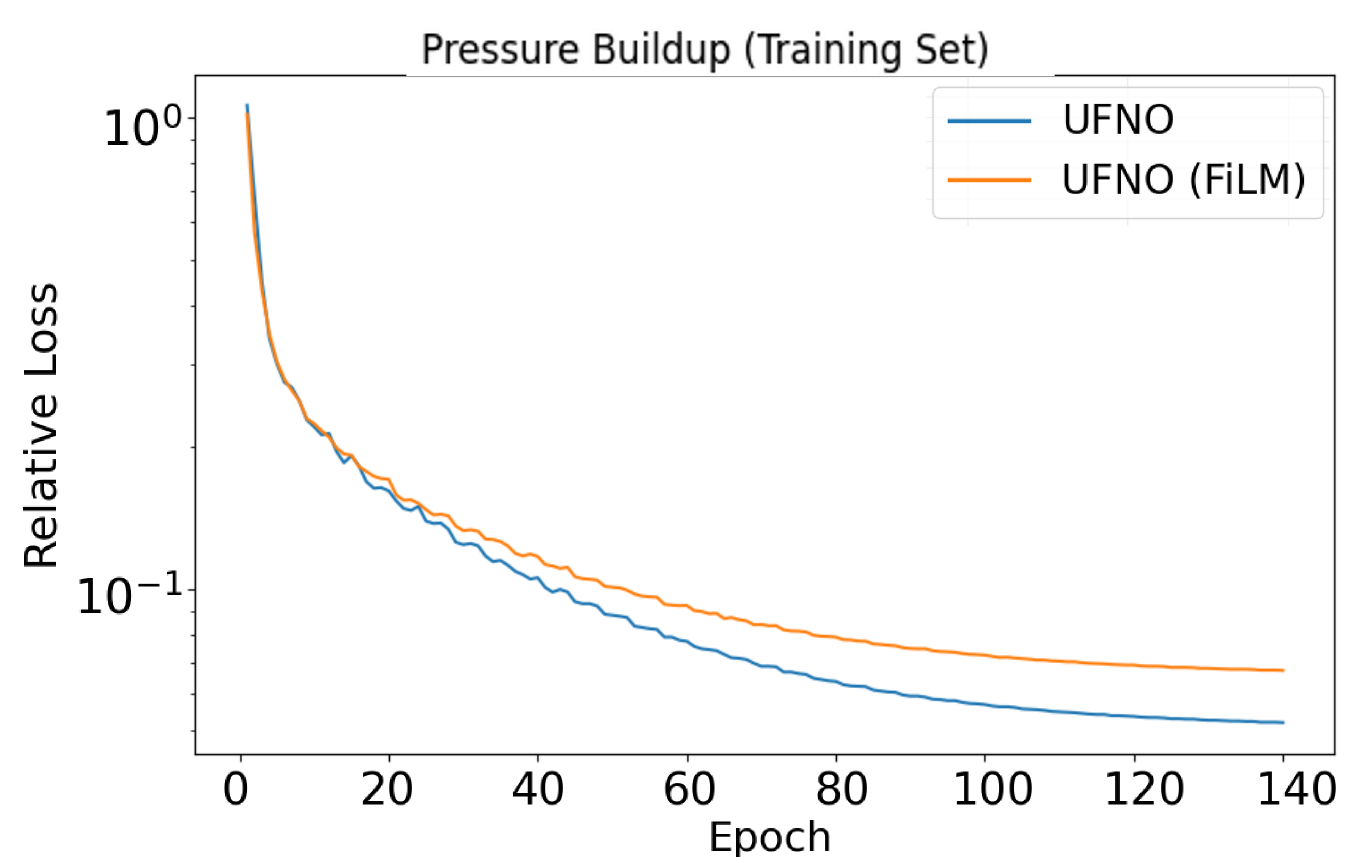} 
        \caption{Training curves}
        \label{fig:training_curves_dp}
    \end{subfigure}
    \hfill
    \begin{subfigure}[b]{0.48\textwidth}
        \centering
        \includegraphics[width=\textwidth]{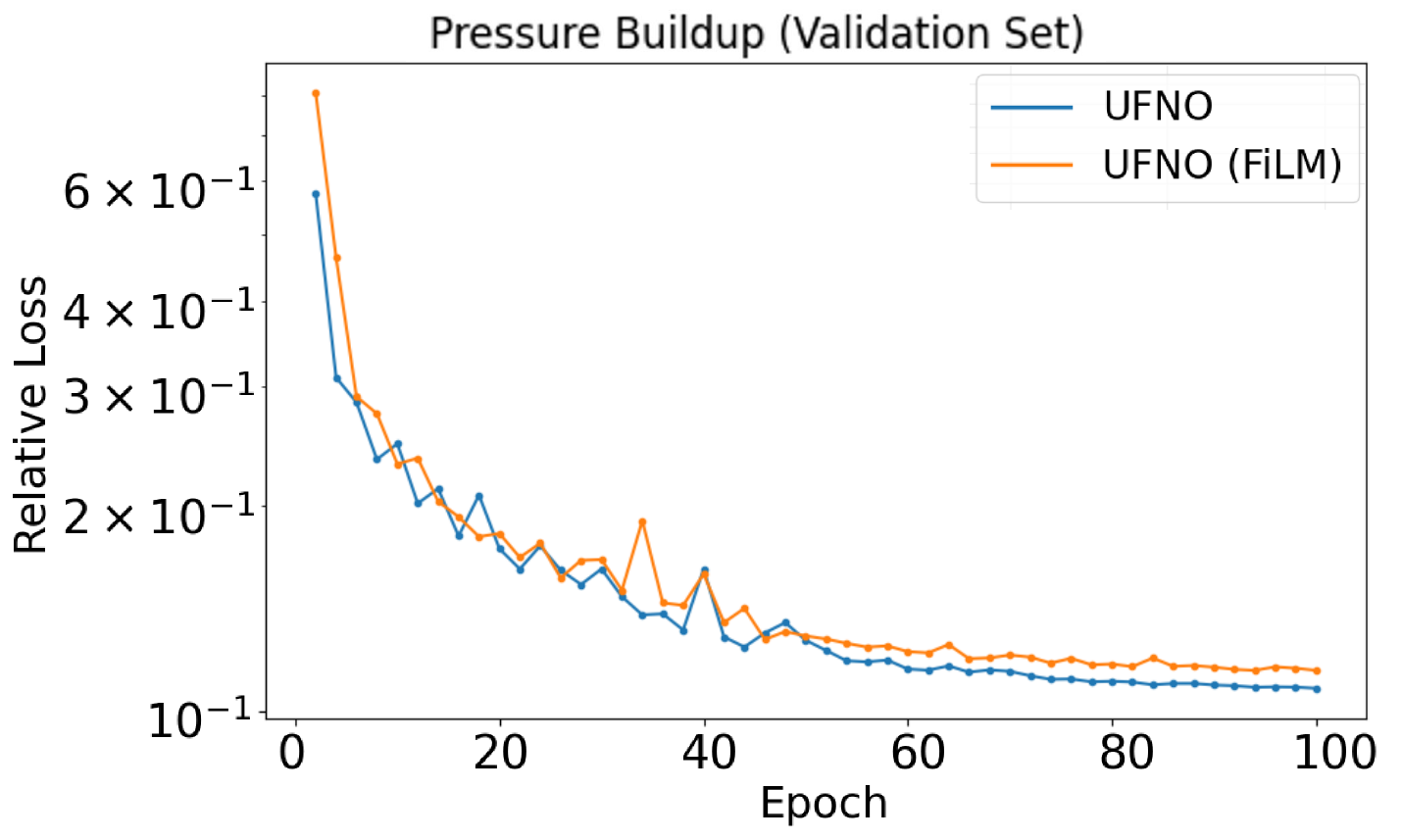}
        \caption{Validation curves}
        \label{fig:validation_curves_dp}
    \end{subfigure}
    \caption{Comparison of training and validation curves for UFNO, and UFNO-FiLM on the pressure build-up problem.}
    \label{fig:curves_side_by_side_dp}
\end{figure}

\paragraph{Pressure build-up results} 
Figure~\ref{fig:curves_side_by_side_dp} shows the relative loss for UFNO and UFNO-FiLM on the training and validation sets for the pressure buildup prediction task. In contrast to the saturation experiments, UFNO achieves lower losses than UFNO-FiLM on both sets, indicating a more efficient fit to the pressure data. The gap in validation performance, however, remains relatively small, suggesting that although FiLM does not provide a clear advantage in this setting, it also does not substantially degrade predictive capability. One possible explanation for this behavior is that pressure fields exhibit smoother spatial variations and stronger global coupling than saturation fields. Because the FiLM mechanism primarily modulates channel-wise spatial features using scalar inputs, it may offer limited benefit when scalar and spatial components are already tightly correlated, as is the case for pressure buildup. Additionally, the spatially weighted loss emphasizes regions with strong saturation dynamics, which may not align with the comparatively diffuse structure of pressure fields, leading to less effective learning for this target variable.

\section{Conclusion}
\label{conclusion}

In this paper, we introduced UFNO-FiLM, a surrogate modeling framework that extends the UNet Fourier Neural Operator (UFNO) through two principal enhancements: a Feature-wise Linear Modulation (FiLM) layer and a spatially weighted loss function. The FiLM layer decouples scalar inputs from spatial features, enabling more expressive and physically meaningful modulation of feature maps while avoiding the redundant transformation of constant signals into the frequency domain. The spatially weighted loss further guides the model to prioritize regions of greater physical relevance, improving its ability to capture complex flow patterns.

Our experimental results demonstrate that UFNO-FiLM achieves consistent improvements over existing neural-operator baselines, including a 21\% reduction in validation error relative to UFNO. These gains are reflected across multiple metrics: Mean Absolute Error (MAE), Structural Similarity Index Measure (SSIM), and Relative Error, s well as in qualitative error visualizations. The findings highlight the value of incorporating FiLM-based conditioning and spatially informed weighting into operator-learning architectures for subsurface flow prediction. While the method shows clear gains for gas saturation, its benefits did not extend to the smoother pressure fields in this setup, suggesting the method is most impactful for problems with high-frequency spatial dynamics Future work will explore extending these components to more complex multiphase systems, three-dimensional domains, and large-scale reservoir simulation settings.

	%\section*{Code Availability}

	\section*{Acknowledgements}
    This work is funded by the Engineering and Physical Sciences Research Council's ECO-AI Project grant (reference number EP/Y006143/1), with additional financial support from the PETRONAS Center of Excellence in Subsurface Engineering and Energy Transition (PACESET).
	\section*{Competing interests}
	The authors declare no competing interests.

\bibliographystyle{ieeetr}
\bibliography{ref}

\end{document}